\documentclass[9pt,shortpaper,twoside,web]{ieeecolor}
\usepackage{generic}
\usepackage{cite}
\usepackage{amsmath,amssymb,amsfonts}
\usepackage{algorithmic}
\usepackage[linesnumbered,ruled]{algorithm2e}
\usepackage{graphicx}
\usepackage{graphicx}
\usepackage{textcomp}
\usepackage{multirow}
\usepackage{mathrsfs}

\usepackage[mathscr]{euscript}

\begin{document}
\title{DETCID: Detection of Elongated Touching Cells with \\ Inhomogeneous Illumination using a Deep Adversarial Network}

\author{Ali Memariani and Ioannis A. Kakadiaris, \IEEEmembership{Member, IEEE}
\thanks{This work was supported in part by the Hugh Roy and Lillie Cranz Cullen Endowment Fund. Ali Memariani and Ioannis A. Kakadiaris are affiliated with the Computational Biomedicine Lab, Department of Computer Science, University of Houston, Houston TX (e-mail: \{amemaria, ikakadia\}@central.uh.edu). 
}
\thanks{We would like to thank Dr. Bradley T. Endress and Dr. Eugenie Basseres for correcting the primary set of annotations and Dr. Jahangir Alam for providing the second set of annotations.}
}

\maketitle

\begin{abstract}
\textit{Clostridioides} difficile infection (C. diff) is the most common cause of death due to secondary infection in hospital patients in the United States. 
Detection of C. diff cells in scanning electron microscopy (SEM) images is an important task to quantify the efficacy of the under-development treatments. 
However, detecting C. diff cells in SEM images is a challenging problem due to the presence of inhomogeneous illumination and occlusion. An Illumination normalization pre-processing step destroys the texture and adds noise to the image. 
Furthermore, cells are often clustered together resulting in touching cells and occlusion. 
In this paper, DETCID, a deep cell detection method using adversarial training, specifically robust to inhomogeneous illumination and occlusion, is proposed. An adversarial network is developed to provide region proposals and pass the proposals to a feature extraction network.
Furthermore, a modified IoU metric is developed to allow the detection of touching cells in various orientations.
The results indicate that DETCID outperforms the state-of-the-art in detection of touching cells in SEM images by at least 20 percent improvement of mean average precision.
\end{abstract}


\section{Introduction}
\begin{figure}[bt]
  \begin{center}
\begin{tabular}{cc} \\
   \includegraphics[width=4cm,height=3cm]{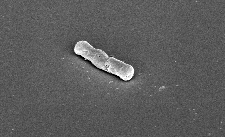}  &  
   \includegraphics[width=4cm,height=3cm]{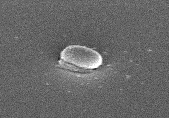}  \\ 
   (a) & (b) \\
   \includegraphics[width=4cm,height=3cm]{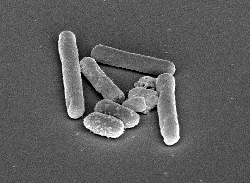}  &  
   \includegraphics[width=4cm,height=3cm]{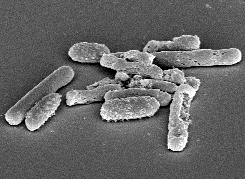}   
\\

(c) & (d)\\
   
   \hspace{-0.55in}  

\end{tabular}
\end{center}
  \caption{Depiction of samples of C. diff cells with challenges of Inhomogeneous illumination and touching cells.
Inhomogeneous illumination makes the segmentation of scanning electron microscopy images challenging: 
(a) shadows on the cell body, 
(b) shadows in the periphery,  
(c) touching cells with overlapping cell walls, and
(d) touching cells with occluded cell bodies.
}
\label{fig:challenges}
\end{figure}

\textit{Clostridioides} difficile infection (C. diff) is the most common cause of death due to infectious gastroenteritis in the USA and a significant source of morbidity \cite{endres2019epidemic}.
Phenotypic features (e.g., length, shape deformation) of C. diff cells in scanning electron microscopy (SEM) images indicate critical information about cell health in C. diff research studies. Therefore, instance-level segmentation of the C. diff cells is an important task to extract phenotypic information \cite{Endres_2016_17411}. 
However, analysis of SEM images is challenging due to inhomogeneous illumination and presence of touching cells. Figure \ref{fig:challenges} depicts samples of C. diff cells in SEM images. 
In computer vision, the terms detection, semantic segmentation, and instance-level segmentation are defined differently. Nevertheless, in microscopy image analysis, these terms are often used interchangeably referring to the latter where the task is to provide a mask for every cell instance in the image.

Classical illumination normalization techniques could reduce the effect of illumination as a pre-processing step. However, they destroy the texture and add noise to the image \cite{Han_2013_14301,MemarianiDeTEC2016}. 
Furthermore, SEM images are limited in number and expensive to obtain. Deep learning methods have achieved state-of-the-art performance in many computer vision tasks such as object detection and segmentation \cite{shen2017deep}. However, deep methods require a large number of training samples. Therefore, preparing a large training set that includes all possible variation of cells is an important task in cell segmentation. 
Exiting data augmentation methods (e.g., flipping, rotation, and warping the entire image) are limited in creating diverse training samples. In current practice, increasing the training samples by an order of magnitude 
results in a highly correlated training set. 


\begin{figure*}[hbt]
  \begin{center}

   \includegraphics[width=15cm]{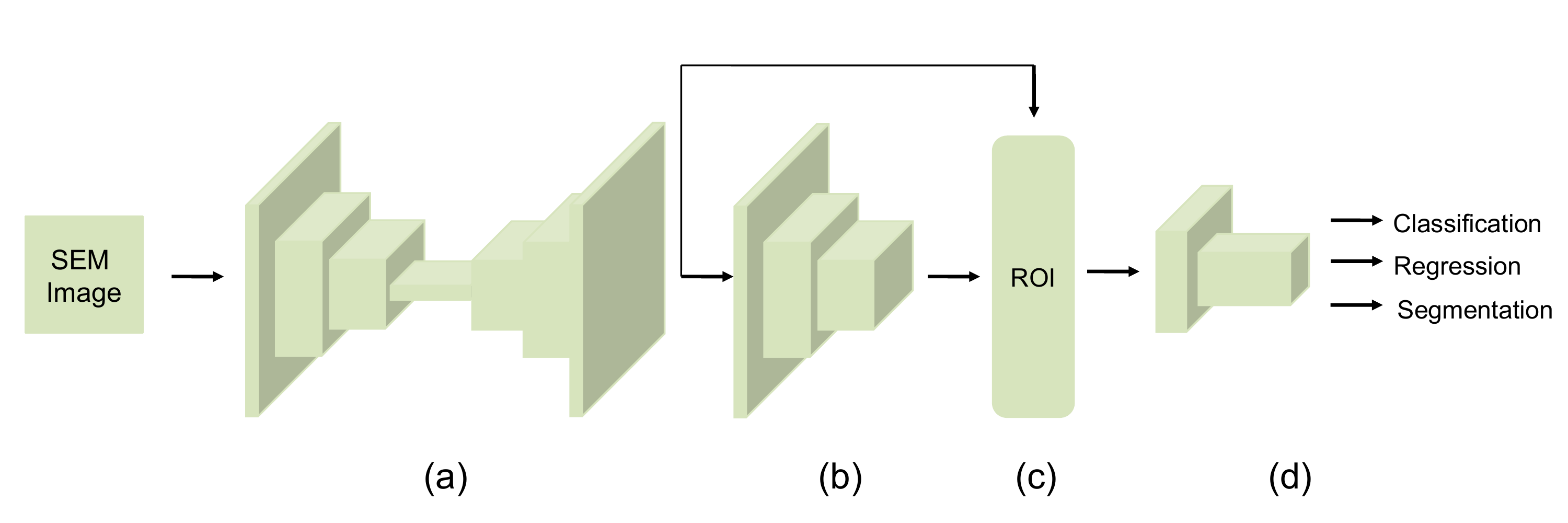}  \\

\end{center}
  \caption{Depiction of DETCID Pipeline: 
  (a) An adversarial region proposal network (ARPN) selects the cell region to be classified, 
(b) an FEN extracts the features from cell candidate regions,
(c) detected region of interests (ROI) are aligned using bilinear interpolation, and
(d) a network head is applied to produce the final mask.}
\label{fig:pipeline}
\end{figure*}

In this paper, we present DETCID (Detection of Elongated Touching C. diff Cells in the presence of Inhomogeneous illumInation using a Deep adversarial network) a deep cell detection algorithm to detect C. diff cells in SEM images. DETCID models the inhomogeneous illumination as an adversarial attack. 
Our previous work, SoLid \cite{memariani2018solid} demonstrated that adversarial training could improve the semantic segmentation performance of U-net in presence of inhomogeneous illumination by 44\% where the adversarial loss penalizes the segmentation output to be similar to the ground truth without increasing the complexity of the network during deployment. 
DETCID improves the state-of-the-art with the following contributions: 
\begin{enumerate}
    \item Developed an adversarial region proposal network to provide candidate bounding boxes for cell detection, reducing the effect of inhomogeneous illumination (Section III.A)
    
    \item Developed a novel non-max suppression method based on mask overlaps to detect touching cells in various orientations (Section III.B)
  
    \item Developed an image synthesis algorithm to generate a large training set of microscopy images for training deep networks (Section III.C).
\end{enumerate}

DETCID expands the semantic segmentation method in SoLiD to an instance-level segmentation method where several bounding boxes are selected to extract features relevant features from a deep feature pyramid network such as ResNet50. Furthermore, the mask-based non-max suppression method detects the clusters of touching cells in various orientations. An image synthesis algorithm is developed to generate clustered cell images to train the network.

The rest of the paper is organized as follows: 
Section II presents a review of the literature related to instance-level cell segmentation.
Section III describes DETCID algorithm. Section IV presents the experimental results comparing the segmentation performance of DETCID with the state-of-the-art. Finally, Section V draws the conclusions.

\section{Related Work}
This section categorizes the related work on cell detection into shallow and deep detection methods. Then, presents the findings from a review of the literature on deep segmentation methods addressing inhomogeneous illumination.

\subsection{Shallow cell detection methods}
Shallow methods select cell candidate regions using intensity thresholding or energy minimization. Then, an optimization algorithm is applied \cite{Arteta_2016_16806, 7784750}, a machine learning regressor is trained over a set of hand-crafted features to select the cell from a set of candidates \cite{Kainz_2015_16488, SantamariaPang_2015_17257, funke2015learning, ahmad2018correlation}, or a level set segmentation is used \cite{Zhou2017Correntropy,zhi2018saliency}.
Finally, size filtering or hole filling is applied to refine the results.
To address the inhomogeneous illumination, pre-processing steps are proposed. However, these pre-processing steps may remove texture information, making the detection more challenging \cite{memariani2018detcic}. 
Ulman \textit{et. al.} \cite{ulman2017objective} performed an objective comparison of many shallow cell detection algorithms with deep convolutional networks.
Shallow methods do not require expensive computing hardware to train and are interpretable. However, they are outperformed by deep learning methods.

\subsection{Deep cell detection methods}
Deep learning algorithms have outperformed the state of the art in many biomedical image processing tasks \cite{shen2017deep}. Shi \textit{et. al.} \cite{shi2019quaternion} applied a cascade of Quaternion Grassmann average layers to develop an unsupervised deep network for segmentation of histology cells. Others have applied deep auto-encoders for cell segmentation \cite{hou2019sparse}. Shen \textit{et. al.} \cite{shen2017deep} reviewed many unsupervised, or CNN-based approaches combined with hand-crafted features for segmentation of biomedical images. 
Roopa \textit{et. al.} \cite{hegde2019comparison} trained a CNN with hand-craft features as input to classify white blood cells in peripheral blood smear images. Hand-crafted cell nuclei boundary masks are also used as shape prior to filter the detection of CNNs \cite{tofighi2018deep}.
Others applied CNNs for cell detection with pixel-level classification for each patch in the images\cite{xue2016cell, xie2018microscopy, tiwari2018detection}. Hofener \textit{et. al.} \cite{hofener2018deep} applied post-processing to smooth the scores derived by CNNs to improve the cell nuclei detection in histology images.
However, patch-based approaches need to run the network for every patch resulting in redundant computations. 

To reduce the computations by sharing the computations over the overlapping patches, fully convolutional networks were introduced for image segmentation  \cite{long2015fully}. Specifically, U-net is widely used for biomedical image segmentation tasks \cite{Ronneberger_2015_17823}. 
Xie \textit{et. al.} \cite{xie2018efficient} evaluated the performance of U-net on multiple pathology datasets.
Ramesh \textit{et. al.} \cite{Ramesh2018cell} added an unsupervised pre-processing layer with logistic sigmoid functions to U-net to separate clustered image patches from each other. 
However, U-net is sensitive to inhomogeneous illumination which increases the false positive for segmentation of SEM images \cite{memariani2018solid}.
Xie \textit{et. al.} \cite{xie2018microscopy} applied two U-shape network architecture without skip connections to compute cell spatial density maps. Spatial densities are applied to detect cell overlaps. However, the application is limited to round shape objects only \cite{xie2018microscopy, naylor2018segmentation}.

Gu \textit{et. al. } \cite{gu2018multi} proposed a pyramid  of residual blocks to capture spatial information in multiple resolutions to detect histology cells in various sizes.
Li \textit{et. al.} \cite{li2018path} replaced the head layers in Mask R-CNN with a conditional random field (CRF) to impose smoothness on the boundaries of segmented patches.

\subsection{Deep methods addressing inhomogeneous illumination}
Deep networks, such as U-net, are sensitive to inhomogeneous illumination. Wan \textit{et. al.} \cite{wan2019simultaneous} proposed an iterative process where a U-net is applied to provide a preliminary segmentation followed by a convolution layer to estimate the bias field in magnetic resonance (MR) images. Next, the bias field corrected image is again sent to the U-net for the next iteration, improving the segmentation. However, the iterative process involves many passes through the network, increasing the complexity of the method.

Generative adversarial networks (GANs) have been used to add robustness to adversarial attacks to the deep networks \cite{goodfellow2014generative}. 
Adversarial training can improve image segmentation by producing label maps that are similar to a target image \cite{luc2016semantic}. 
Adversarial networks have been applied in the segmentation of MR images \cite{li2017brain,moeskops2017adversarial,xue2017segan} where the datasets and the annotations are available. However, the application is limited since the adversarial training requires a large training set to train both the segmenter and the discriminator networks.
Lee \textit{et. al.} \cite{lee2019three} proposed an unsupervised image deconvolution method using a cycle-consistent adversarial network to improve the quality of blurred and noisy fluorescence microscopy images without labeled data. The adversarial network in DETCID models the inllumination as an adversarial attack without during training without increasing the complexity of the network during deployment.

\section{Methods}
DETCID compromises of two parts: a deep adversarial region proposal network (ARPN) and a feature extraction network (FEN). Figure \ref{fig:pipeline} depicts the overview of the pipeline.
The input to ARPN is a cell image and the output is a label map. The FEN is fully convolutional and produces a probability map for presence of cells in the image. 
An RoI alignment layer combines the output of the two network and aligns the extracted features with the input. RoIs are passed through two convolution layers to produce the final segmentation mask.

\subsection{Adversarial region proposals network}
The ARPN consists of two deep ConvNets, namely the segmenter and the discriminator. The segmenter predicts a label map for the pixels while the discriminator distinguishes between the predicted label maps and the ground truth. 

The input to the segmenter is a cell image. The segmenter includes a convolution path and a deconvolution path similar to U-net. The convolution path extracts a feature map for segmentation using convolution layers while the deconvolution path increases the resolution, creating a label map. 
The generated label map may differ significantly from the ground truth since the segmenter does not consider the smoothness of the labels, resulting in a non-continuous segmentation.

A second ConvNet (discriminator) is used to train the segmenter to produce label maps similar to the ground truth.
The discriminator is a regular ConvNet classifier trained on the ground truth and predicted segmentations. 
During training, it learns to classify the input image into two classes: ``artificially generated" or ``ground truth", and backpropagates the gradients.

\subsubsection{Segmenter Network}
The segmenter network consists of six convolutional units: the first three units include a 3x3 convolution layer, a ReLU layer and a 2x2 max pooling layer with a stride of two downsampling the image (contracting units). 
The next three units (expanding units) include an upsampling of the features followed by a 2x2 deconvolution. 
Each contracting unit doubles the number of feature channels while each expanding unit halves the number of channels.
The segmenter minimizes a loss function $L_S$:
\begin{equation}
L_S = \mathbf{w_c} * L_C \Big( \mathcal{S}(\mathbf{I}), \mathbf{G} \Big) + L_C\Big( \mathcal{D}\big(\mathcal{S}(\mathbf{I})\big), 1 \Big),
\label{eq:segmenterLoss}
\end{equation}
where $L_C \Big( \mathcal{S}(\mathbf{I}), \mathbf{G} \Big)$ is a cross-entropy term between the predicted labels $S$ corresponding to the image $\mathbf{I}$ and the ground truth $\mathbf{G}$. The second term $L_C\Big( \mathcal{D}\big(\mathcal{S}(\mathbf{I})\big), 1 \Big)$ is the  adversarial loss term, computed by the discriminator. 
The label map of image $\mathbf{I}$ generated by the segmenter is denoted by $\mathcal{S}(\mathbf{I})$ and $\mathcal{D}$ is the discriminator network described in the next section. The adversarial loss forces the segmenter to produce label maps that would be considered as ground truth by the discriminator.
To distinguish touching cells, the segmenter considers the boundaries of the cells (cell wall) as a separate class. Hence, the segmenter loss is one-hot encoded with three classes. The number of cell wall samples are considerably less compared to the other classes. To compensate for the bias in the training set, the segmenter cross-entropy loss is weighted ($\mathbf{w_c}$). The minority class receives a higher classification weight.

The segmenter network may misclassify a large portion of cells due to inhomogeneous illumination. The adversarial training is used to evaluate such misclassifications and improve the segmenter. A discriminator ConvNet is used to compute the likelihood of the predicted segmentation map being an actual label map.

\subsubsection{Discriminator Network}
The discriminator improves the generated labels by sending feedback to the generator if the segmentation labels are significantly different from the ground truth. It does not increase the complexity of the network since it is used only during training.
It consists of five convolutional layers with valid padding, followed by ReLU activations and average pooling. Furthermore, two fully connected layers are placed at the end of the discriminator. 

To avoid saturation, the last layer of the discriminator does not have a thresholding operator so it produces an unscaled output. Computing scores between 0 and 1 may cause the discriminator to generate values close to 0 for generated label maps, in which case the gradient would be too small to update the generator and eventually saturate the network \cite{goodfellow2014generative}. 

The discriminator ($\mathcal{D}$) computes the cross-entropy of the ground truth label maps ($\mathbf{G}$) and 1, and the cross-entropy of the generated label maps ($S(\mathbf{I})$) and 0, minimizing the following loss function:

\begin{equation}
L_D = L_C \Big( \mathcal{D}(\mathbf{G}),1 \Big)+ L_C\Big( \mathcal{D}(\mathcal{S}(\mathbf{I})), 0 \Big).
\label{eq:discriminatorLoss}
\end{equation}

During the training, the discriminator improves the segmenter network, penalizing the segmentation labels that do not look like manual labels.
Therefore, the adversarial result has properties such as smoothness and robustness to inhomogeneous illumination.

\begin{figure*}[htb]
  \begin{center}
\begin{tabular}{ccc} \\
    \hspace{-0.18in}
   \includegraphics[width=6cm]{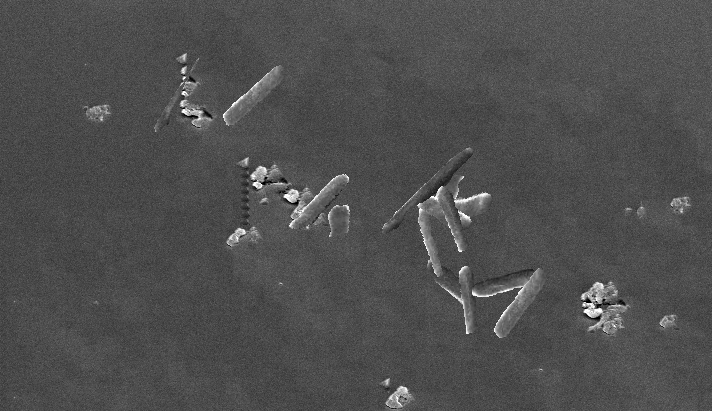}
    &
     \hspace{-0.18in}
  \includegraphics[width=6cm]{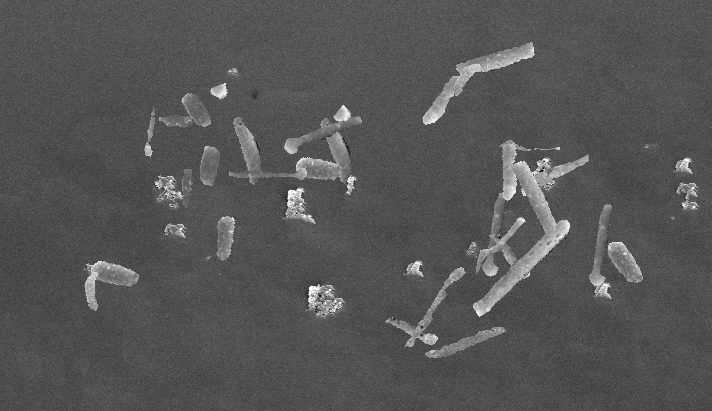}
  &
    \hspace{-0.18in}
  \includegraphics[width=6cm]{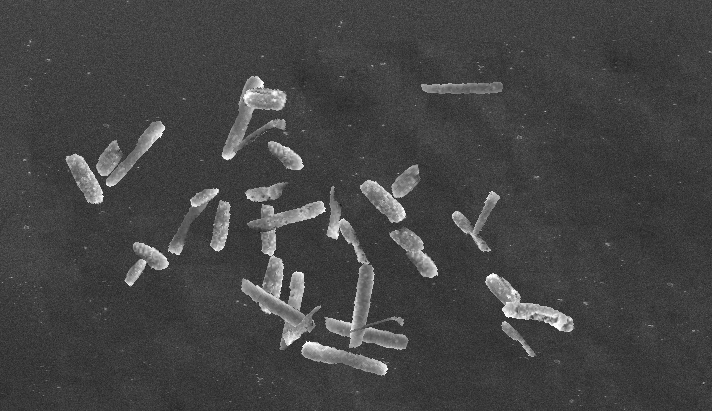}  \\
    \hspace{-0.18in}
  \includegraphics[width=6cm]{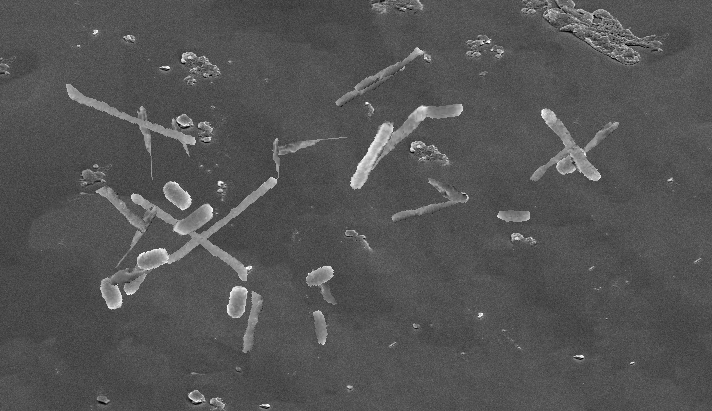} &
  \hspace{-0.18in}
  \includegraphics[width=6cm]{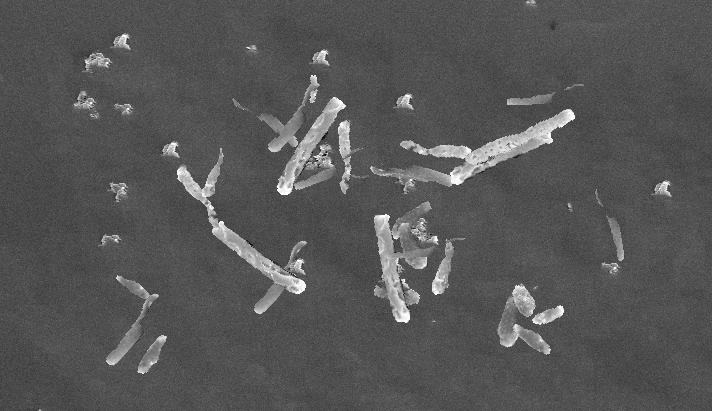} &
  \hspace{-0.18in}
    \includegraphics[width=6cm]{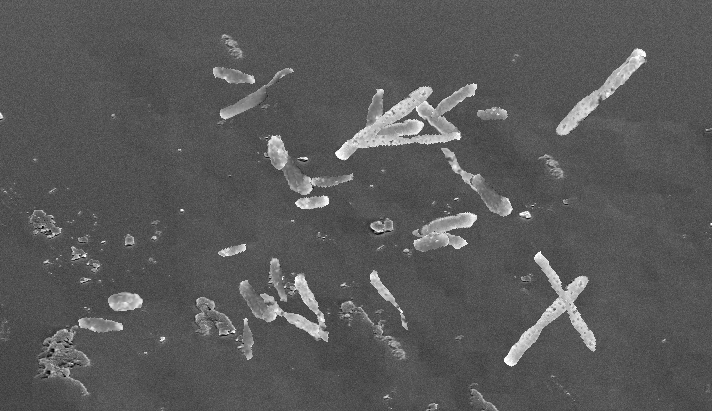} \\
    \hspace{-0.18in}

\end{tabular}
\end{center}
  \caption{ (T) Depiction of selected the synthesized SEM images of overlapping C. diff cells with inhomogeneous illumination and (B) with the presence of debris.}
\label{fig:synthetic}
\end{figure*}

\subsection{Instance-level cell segmentation}
A ResNet50 architecture \cite{Lin_2017_CVPR} is applied to extract features from images. The computation is shared for all cell bounding boxes for efficiency. 
Generating a large number of bounding box proposals is one of the major drawbacks of algorithms such as MaskRCNN \cite{He_2017_ICCV} and Faster R-CNN \cite{NIPS2015_5638}. Even with feature sharing using ResNet, region based methods computational complexity is not comparable with one shot detection algorithms such as YOLO \cite{Redmon_2016_CVPR}.
Unlike Mask R-CNN, DETCID uses the APRN to generate the proposal bounding boxes to reduce the number of the proposals. An average pooling is applied on the result of the ARPN for each anchor type (i.e., horizontal, vertical, and square box). If the average value is greater than a threshold \textit{t}, then anchor boxes are centered at that location. The values of the threshold is determine by the size of the smallest cell in the training set. 

Manual annotation assigns a single bounding box to a cell. However, small variations of the bounding box in length and height may also get a high classification score by the detection method due to smoothness property. Setting all such variations to the background will be confusing for the network. 
Furthermore, the region proposal network may not always propose the finest bounding box around the object. 
Therefore, for each proposed region, similar bounding boxes with variations in length and height are passed to a fully-connected layer for refinement.  

ResNet features are extracted for each anchor box and are fed to the ROI alignment. 
A network head is applied similar to head in Mask-RCNN to compute the masks, bounding boxes, and detection probabilities. The cells may appear in various orientations. Therefore, applying non-max suppression based on the bounding box overlaps is not suitable for detection of cells in SEM images. DETCID modifies the computation of intersection over union (IoU) based on the masks overlaps and the area of the cell masks. Non-max suppression is then applied on the modified IoU values. 

The loss function includes a classification term to detect the cells and a regression term to identify the bounding box:

\begin{equation}
L_F = L_C ( p,p^*) + p^*L_R( \mathbf{r},\mathbf{r^*} )+ L_C(\mathbf{M},\mathbf{G}),
\label{eq:detectionLoss}
\end{equation}
where, $ p$ and $p^*$ denote the classification score and label respectively, $r,r^*$ denote the prediction and annotated bounding box parameters, and $\mathbf{M}$ and $\mathbf{G}$ are the predicted and the ground truth masks respectively labeling the pixels as background or cell candidate.

\begin{table*}[bt]
\caption{Definition of notations used in Algorithm 1. }
\centering

\begin{tabular}{ | c | c | l | }
\hline
	Notation & Type & Definition \\ \hline
	$\phi$ & Scalar & Angle between the major axes of a crossing pair of cells (degrees) \\ \hline
	$\theta$ & Scalar & Angle of the major axis of the cell (degrees) \\ \hline
	$(x,y)$ & Scalar & Centroid of a cell (pixels) \\ \hline
	$e,f$ & Scalar & Constants \\ \hline
	$\eta$ & Scalar & Height of a cell mask (pixels) \\ \hline
	$\sigma$ & Scalar & Horizontal shear parameter \\ \hline
	$(\rho,\kappa)$ & Scalar & Image size for acquired and synthetic images (pixels) \\ \hline
	
	$\delta$ & Scalar & Maximum change of the random orientation of a cell (degrees) \\ \hline
	$\chi$ & Scalar & Maximum horizontal shift of the second cell in a touching/crossing pair (pixels) \\ \hline
	$\psi$ & Scalar & Maximum vertical shift of the second cell in a touching/crossing pair (pixels)\\ \hline
	$a$ & Scalar & Number of acquired images \\ \hline
	$n$ & Scalar & Number of cells in an acquired image \\ \hline
	$c$ & Scalar & Number of crossing pairs of cells in a synthetic image \\ \hline
	$o$ & Scalar & Number of isolated cells in a synthetic image \\ \hline
	$t$ & Scalar & Number of touching pairs of cells in a synthetic image \\ \hline
	$z$ & Scalar & Number of synthetic images \\ \hline
	
	$\epsilon$ & Scalar & Random number in the range [0,1] \\ \hline
	$\omega$ & Scalar & Width of a cell mask (pixels) \\ \hline
	$w$ & Scalar & Window size for texture synthesis (same width and height) (pixels) \\ \hline
	
	$\mathbf{A}^{k}$ & 2D tensor & Acquired image \textit{k} \\ \hline
	\rule{0pt}{2ex} $\mathbf{B}^{k}_i$ & 2D tensor & Annotation mask for image \textit{k} and cell \textit{i} with size $(\rho, \kappa)$ \\ \hline
	\rule{0pt}{2.5ex} $\mathbf{J}^{l}_j$ & 2D tensor & Synthetic ground truth mask for image \textit{l} and cell \textit{j} with size $(\rho, \kappa)$ (pixels)\\ \hline
	$\mathbf{C}$ & 2D tensor & Image of a cell with size $(\rho, \kappa)$ (pixels)\\ \hline
	$\mathbf{C}^m$ & 2D tensor & Binary mask of a cell with size $(\rho, \kappa)$ (pixels)\\ \hline
	$\mathbf{I}^{l}$ & 2D tensor & The \textit{l}\textsuperscript{th} synthetic image  \\ \hline
	$\textbf{T}$ & 2D tensor & 3$\times$3 geometric transformation matrix \\ \hline
	
	$\mathbf{B}^{k}$ & 3D tensor & Annotation mask for acquired image \textit{k} with size $(n, \rho, \kappa)$ (pixels) \\ \hline
	$\mathbf{J}^{l}$ & 3D tensor & Synthetic ground truth mask for image \textit{l} with size $(o+2t+2c, \rho, \kappa)$ (pixels)\\ \hline
	
	$\pmb{\mathscr{A}}^m$ & Set & Set of manually annotated masks \\ \hline
	$\pmb{\mathscr{A}}^r$ & Set & Set of acquired images \\ \hline
	$\pmb{\mathscr{I}}^g$ & Set & Set of synthetic ground truth masks \\ \hline
	$\pmb{\mathscr{I}}^s$ & Set & Set of synthetic cell images \\ \hline
	
\end{tabular}
\label{fig:notations}
\end{table*}

\begin{algorithm*}[htb]
    \SetKwInOut{Input}{Input}
    
    \SetKwInOut{Output}{Output}
    
    \Input{A set of $a$ acquired images $\pmb{\mathscr{A}}^r = \{\mathbf{A}^{1}, ..., \mathbf{A}^{a} \}$ and their manually annotated masks $\pmb{\mathscr{A}}^m = \{\mathbf{B}^{1}, ..., \mathbf{B}^{a} \}$, number of images to be synthesized $z$, window size $m$}
    \Output{Synthetic cell images $\pmb{\mathscr{I}} = \{\mathbf{I}^{1}, ..., \mathbf{I}^{z} \}$, synthetic ground truth masks $\pmb{\mathscr{J}} = \{\mathbf{J}^{1}, ..., \mathbf{I}^{z} \}$}    
    
    \SetKwFunction{FOverlay}{AddCell}
    \SetKwProg{Fn}{Function}{:}{}
  \Fn{\FOverlay{$\mathbf{A}^{k}$, $\mathbf{B}^{k}_i$, $\mathbf{I}^{l}$, $\mathbf{J}^{l}_j$, $\theta$, $x$, $y$}}{
    \begin{itemize}
        \item{Extract the $i$\textsuperscript{th} cell in the $k$\textsuperscript{th} acquired image $\mathbf{A}^{k}$ using its annotated mask $\mathbf{B}^{k}_i  $: ${\mathbf{C} \leftarrow \mathbf{A}^{k} \circ \mathbf{B}^{k}_i }$, ${\mathbf{C}^m \leftarrow \mathbf{B}^{k}_i }$\; }
        \item{Translate the center of the annotation mask of the cell in $\mathbf{C}$ and its mask $\mathbf{C}^m$ to the center of the image\;}
        \item{Initialize a 3$\times$3 geometric transformation matrix 
        $\textbf{T}$ as an identity matrix and add a random noise to the transformation \newline
        parameters (e.g., horizontal shear: $\sigma = f+e \epsilon$);}
        \item{Warp the masked cell image $\mathbf{C}$ and its annotation mask $\mathbf{C}^m$ with transformation $\textbf{T}$ and resize to size $(\rho, \kappa)$\;}
        \item{Compute the angle $\phi$ of the major axis of the cell corresponding to the horizontal axis\;}
        \item{Rotate $\mathbf{C}$ and $\mathbf{C}^m$ to align with orientation $\theta$\ and crop to the size $(\rho, \kappa)$;}
        \item{Translate the object centroid in $\mathbf{C}$ and $\mathbf{C}^m$ to location $(x,y)$ (move the object inside if the cell is partially outside the image boundaries);}
        \item{Overlay $\mathbf{C}$ on $\mathbf{I}^{l}$ and $\mathbf{J}^{l}_j \leftarrow 
        \mathbf{C}^m$\;}
    \end{itemize}
        \KwRet $\mathbf{I}^{l}$, $\mathbf{J}^{l}_j$\;
  }

    \Begin(){   
    \For{$l=1, ..., z$}{
    	\begin{itemize}
        \item{Randomly select an image $\mathbf{A}^{k}$ from the acquired images $\pmb{\mathscr{A}}^r$ and its annotation mask $\mathbf{B}^{k}$ from 
        $\pmb{\mathscr{A}}^m$\;}
        \item{Apply the image inpainting algorithm \cite{korman2015coherency} to remove the cells from the image and store the background to $\mathbf{I}^{l}$\;}
        \item{Randomly select a background patch of size $w \times w$ 
        from image $\mathbf{I}^{l}$\; }
        \item{Synthesize a background image with the same 
        resolution as $\mathbf{I}^{l}$ using the texture synthesis 
        algorithm \cite{Efros_2001_17825} and \newline
        replace with $\mathbf{I}^{l}$\;}

    	\item{Randomly select the number of isolated cells $o$, touching pairs $t$, 
    	and crossing pairs $c$ to be placed into \newline
    	the image ($o,t,c \in \{1,...,n\}$)\;}
    	\item{Initialize $\mathbf{J}^{l}$ with zeros as a 3D tensor of size ($o+2t+2c, \rho, \kappa$)}
	\end{itemize}
	
	\For{$p=1, ..., o$}{
	\begin{itemize}
    	\item{Randomly select a cell mask $i$ ($i \in {1,...,n}$) with annotation tensor $\mathbf{B}^{k}_i$ in the acquired image $\mathbf{A}^{k}$\;}
    	\item{Randomly generate $\theta \in [0^{\circ},180^{\circ}]$, $x_1 \in [\frac{\omega}{2}, ..., \rho-\frac{\omega}{2}]$, $y_1 \in [\frac{\eta}{2}, ..., \kappa-\frac{\eta}{2}]$\;}
    	\item{Add the cell into the image: $\mathbf{I}^{l}$, $\mathbf{J}^{l}_j \leftarrow$  \FOverlay{$\mathbf{A}^{k}$, $\mathbf{B}^{k}_i$, $\mathbf{I}^{l}$, $\mathbf{J}^{l}_j$, $\theta$, $x_1$, $y_1$}\;}
    	
	\end{itemize}
	}
	\For{$q=1, ..., t$}{
        \begin{itemize}
            \item{Randomly select a cell mask $i$ ($i \in {1,...,n}$) with annotation tensor $\mathbf{B}^{k}_i$ in the acquired image $\mathbf{A}^{k}$\;}
        	\item{Randomly generate $\theta \in [0^{\circ},180^{\circ}]$, $x_1 \in [\frac{\omega}{2}, ..., \rho-\frac{\omega}{2}]$, $y_1 \in [\frac{\eta}{2}, ..., \kappa-\frac{\eta}{2}]$\;}
        	\item{Add the first touching pair into the image the first output: $\mathbf{I}^{l}$, $\mathbf{J}^{l}_j$ $\leftarrow$ \FOverlay{$\mathbf{A}^{k}$, $\mathbf{B}^{k}_i$, $\mathbf{I}^{l}$, $\mathbf{J}^{l}_j$, $\theta$, $x_1$, $y_1$}\;}
        	
        	\item{Update $(x_1, y_1) \leftarrow$ centroid of $\mathbf{J}^{l}_j$\;}
    	   
    	   \item{Randomly select a cell mask $u$ with annotation tensor $\mathbf{B}^{k}_u$ in the acquired image $\mathbf{A}^{k}$ as the second cell\;}
    	   \item{Let $\omega$ be the width and $(x_1,y_1)$ the centroid of the cell mask $\mathbf{C}^m$. Randomly select a location $(x_2, y_2)$: \newline
    	   $x_2 \in \left[x_1-\omega, x_1-\frac{\omega}{2}\right] \cup \left[x_1+\frac{\omega}{2}, x_1+\omega\right] $ and $y_2 \in \left[ y_1- \psi \epsilon , y_1+ \psi \epsilon  \right]$\;}
    	   \item{Add the second touching pair into the image: $\mathbf{I}^{l}$, $\mathbf{J}^{l}_v$ $\leftarrow$ \FOverlay{$\mathbf{A}^{k}$, $\mathbf{B}^{k}_u$, $\mathbf{I}^{l}$, $\mathbf{J}^{l}_v$, $\theta$, $x_2$, $y_2$}\;}
    	   
        \end{itemize}
    }
    \For{$r=1, ..., c$}{
        \begin{itemize}
            \item{Randomly select a cell mask $i$ ($i \in {1,...,n}$) with annotation tensor $\mathbf{B}^{k}_i$ in the acquired image $\mathbf{A}^{k}$\;}
        	\item{Randomly generate $\theta \in [0^{\circ},180^{\circ}]$, $x_1 \in [\frac{\omega}{2}, ..., \rho-\frac{\omega}{2}]$, $y_1 \in [\frac{\eta}{2}, ..., \kappa-\frac{\eta}{2}]$\;}
        	\item{Add the first crossing pair into the image: $\mathbf{I}^{l}$, $\mathbf{J}^{l}_j$ $\leftarrow$ \FOverlay{$\mathbf{A}^{k}$, $\mathbf{B}^{k}_i$, $\mathbf{I}^{l}$, $\mathbf{J}^{l}_j$, $\theta$, $x_1$, $y_1$}\;}
        	
        	\item{Update $(x_1, y_1) \leftarrow$ centroid of $\mathbf{J}^{l}_j$\;}
    	   
    	   \item{Randomly select a cell mask $u$ with annotation tensor $\mathbf{B}^{k}_u$ in the acquired image $\mathbf{A}^{k}$ as the second cell\;}
    	   \item{Let $\eta$ be the height and $(x_1,y_1)$ the centroid of the cell mask $\mathbf{C}^m$. Randomly select a location $(x_2, y_2)$: \newline
    	 $y_2 \in \left[y_1-\eta, y_1-\frac{\eta}{2}\right] \cup \left[y_1+\frac{\eta}{2}, y_1+\eta\right] $ and $x_2 \in \left[ x_1-\chi \epsilon  , x_1+\chi \epsilon   \right]$\;}
    	 \item{Randomly select an angle $\phi$: $\phi=\theta+90 \pm \delta \epsilon  $ }
    	   \item{Add the second crossing pair into the image: $\mathbf{I}^{l}$, $\mathbf{J}^{l}_v$ $\leftarrow$ \FOverlay{$\mathbf{A}^{k}$, $\mathbf{B}^{k}_u$, $\mathbf{I}^{l}$, $\mathbf{J}^{l}_v$, $\phi$, $x_2$, $y_2$}\;}
        \end{itemize}
    }
    
  }

    }         
   \caption{Image synthesis algorithm generating images with isolated, touching, and crossing cells. Notations are defined in Table \ref{fig:notations}}
   \label{fig:augment_algo}
\end{algorithm*}

\subsection{Data Augmentation}
Deep networks require large numbers of training data. Specifically, deep learning architectures proposed for biomedical image analysis are hindered by the lack of large amounts of training data.
Therefore,  generating synthetic images becomes important in the analysis of biomedical images, since their acquisition is expensive and time-consuming. A simple solution would be to avoid training by applying a pre-trained model. However, the solution is limited to cases where a pre-trained network with similar training data exits \cite{shen2017deep}. 
To train a deep segmentation model, Algorithm \ref{fig:augment_algo} is developed which is capable of synthesizing cell images with inhomogeneous illumination where cells could be isolated, touching, or crossing.

Deep learning frameworks such as TensorFlow provide simple image augmentation functions such as translation, rotation, cropping, flipping, and scaling \cite{abadi2016tensorflow}. These basic augmentation techniques often create black areas in the image which could be filled with interpolation or more complex image in-painting techniques.
The basic augmentation methods mentioned above have been previously used to increase the number of training samples and add invariance to challenges such as rotation of objects in the image \cite{dai2017scan,li2017brain, luc2016semantic,moeskops2017adversarial,shen2017deep, van2016deep,xue2017segan}.

Increasing the amount of data by orders of magnitude is essential in training deep models to analyze SEM images. 
However, using basic augmentation methods to increase the amount of data by orders of magnitude results in high correlation between the images in the training dataset.
U-net applied image warping to the cell images, creating images with slightly different cell and backgrounds \cite{Ronneberger_2015_17823}. Nevertheless, warping the whole image with the same warping transformation would limit the number of synthesized images. 

Algorithm \ref{fig:augment_algo} presents an image synthesis algorithm capable of synthesizing large numbers of images with the same background texture and cell shapes in the images captured by SEM.
First, SEM-acquired images are manually annotated to three classes, namely cell body, cell wall, and background. The image quilting technique by Efros and Freeman is applied to synthesize similar background images \cite{Efros_2001_17825}. Then, the cells are randomly warped and placed into the image. Warping the cells ensures that the training data are different from the testing data. Figure \ref{fig:synthetic} depicts samples of synthesized isolated images of isolated, touching, and crossing cells in presence of inhomogeneous illumination and debris.

\subsection{Implementation details}
The implementation of the adversarial region proposal network is based on the previous work by the authors \cite{memariani2018solid} and random patches of size 256$\times$256 are passed to the ResNet50 \cite{Lin_2017_CVPR} for training. The ResNet50 architrave applies scale anchors of size $\{32^2, 64^2, 128^2\}$ to extract features and is initialized with COCO pretrained weights. To allow finer ROIs, five percent variations in length and height of the bounding boxes are considered and bounding boxes with less than five percent of potential cell areas are filtered before pooling ROIs. The above thresholds are found empirically.
Adam optimization is applied to optimize the overall loss function on a cluster with 4 NVIDIA GeForce GTX GPU of 12 GB capacity.

\begin{figure*}[hbt]
  \begin{center}
\begin{tabular}{ccccc} \\
\hspace{-0.1in}
   \includegraphics[width=3.5cm,height=3.5cm]{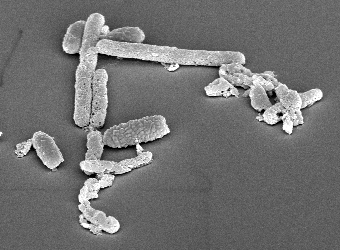} &
   \hspace{-0.18in}
   \includegraphics[width=3.5cm,height=3.5cm]{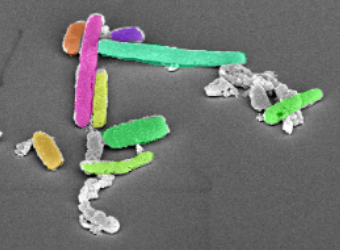} &
  \hspace{-0.18in}
    \includegraphics[width=3.5cm,height=3.5cm]{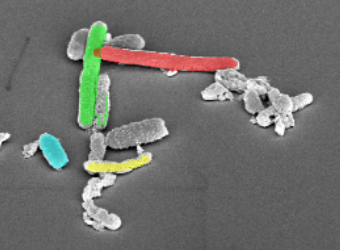} &
    \hspace{-0.18in}
    \includegraphics[width=3.5cm,height=3.5cm]{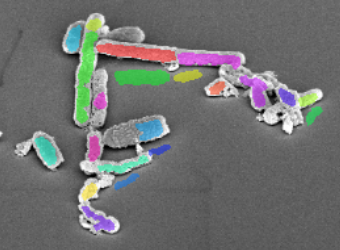} 
    &
    \hspace{-0.18in}
   \includegraphics[width=3.5cm,height=3.5cm]{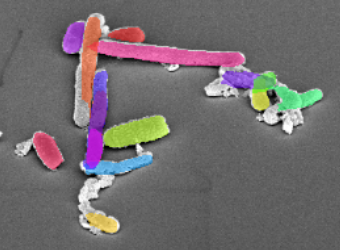}  \\
   
  \hspace{-0.1in}
  \includegraphics[width=3.5cm,height=3.5cm]{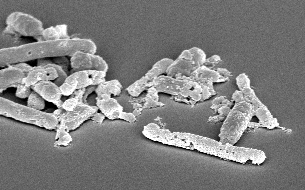} &
  \hspace{-0.18in}
  \includegraphics[width=3.5cm,height=3.5cm]{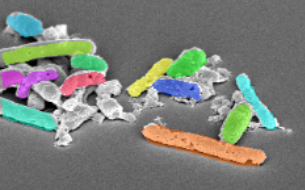} &
  \hspace{-0.18in}
     \includegraphics[width=3.5cm,height=3.5cm]{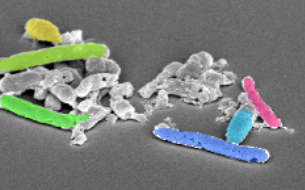} &
     \hspace{-0.18in}
     \includegraphics[width=3.5cm,height=3.5cm]{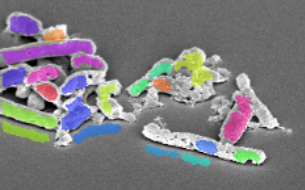}
    &
    \hspace{-0.18in}
  \includegraphics[width=3.5cm,height=3.5cm]{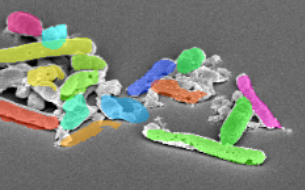}  \\
   
   \hspace{-0.1in}
  \includegraphics[width=3.5cm,height=3.5cm]{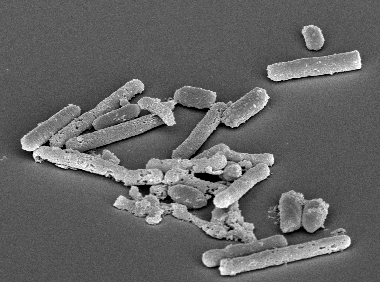} &
  \hspace{-0.18in}
  \includegraphics[width=3.5cm,height=3.5cm]{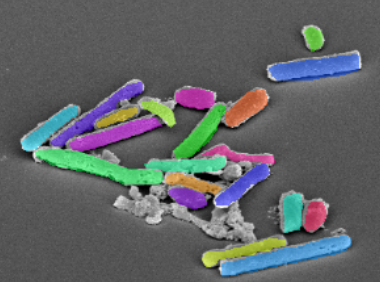} &
  \hspace{-0.18in}
  \includegraphics[width=3.5cm,height=3.5cm]{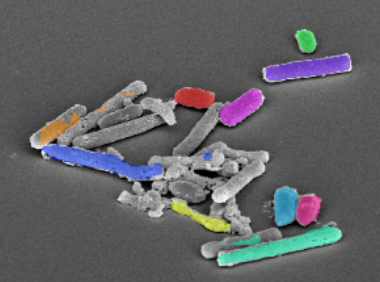} &
  \hspace{-0.18in}
  \includegraphics[width=3.5cm,height=3.5cm]{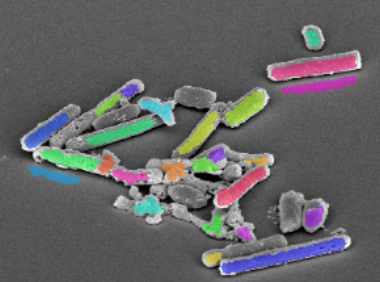}
    &
    \hspace{-0.18in}
  \includegraphics[width=3.5cm,height=3.5cm]{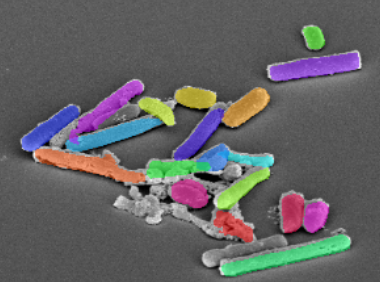}  \\
   
   \hspace{-0.1in}
  \includegraphics[width=3.5cm,height=3.5cm]{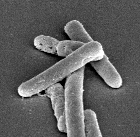} &
  \hspace{-0.18in}
  \includegraphics[width=3.5cm,height=3.5cm]{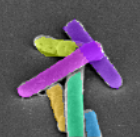} &
  \hspace{-0.18in}
    \includegraphics[width=3.5cm,height=3.5cm]{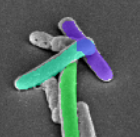} &
    \hspace{-0.18in}
    \includegraphics[width=3.5cm,height=3.5cm]{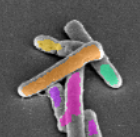}
    &
    \hspace{-0.18in}
  \includegraphics[width=3.5cm,height=3.5cm]{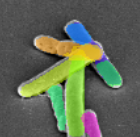}  \\
   
   (a) & (b) & (c) & (d) & (e) \\ 

\end{tabular}
\end{center}
  \caption{Depiction of the segmentation results: (a) original image, (b) ground truth labels, (c) Mask-RCNN, (d) FCRN, and (e) DETCID segmentation. Mask-RCNN is more accurate in detecting isolated cells. However, Mask-RCNN does not detect cells in presence of debris or cells clusters. FCRN is sensitive to inhomogeneous illumination and presence of debris and results in false positives. DETCID is able to detect cells when touching cells are clustered together.}
\label{fig:seg_result}
\end{figure*}

\section{Results and discussion}
To evaluate DETCID for cell detection and segmentation, an acquired SEM dataset UH-A-cdiff1 \cite{memariani2018detcic}, and a synthetic dataset UH-S-cdiff1 were used. Mean average precision (mAP) and dice score metrics were used to evaluate the performance of detection and segmentation, respectively.

\subsection{Datasets}

\subsubsection{UH-A-cdiff1}
UH-A-cdiff1 consists of 22 C. diff cell images (197 vegetative cells and 111 spores) acquired via scanning electron microscopy. Image dimensions are 411$\times$711 pixels with 10,000x magnification. 
Moreover, many cells are touching or crossing each other with the existence of debris. Also, the cells were partially deformed and cell walls are damaged due to a laboratory treatment, making the detection challenging. Two set of annotations were provided labeling every cell as a binary mask for every cell in the image. 

\subsubsection{UH-S-cdiff1}
According to ImageNet \cite{deng2009imagenet} statistics, the rule of thumb for size of training set is 1,000 samples. Algorithm 1 is applied to synthesize 6,000 images of 411$\times$711 pixel dimensions, considering at least 1,000 samples for isolated, touching, and crossing cases of vegetative cells and spores. The number of cases per image is chosen empirically.

The synthetic images include two to four pairs of the following scenarios: a pair of two vegetative cells touching, a pair of two vegetative cells crossing, and a vegetative cell touching a spore. Moreover, two to four single isolated cells of each type is added into the image, creating a variety of possible overlaps between cells in various orientations.

\begin{table*}[bt]
\caption{Quantitative results between the performance of DETCID and
the state-of-the-art in cell detection by Mask-RCNN and FCRN on the acquired (UH-A-cdiff1) and the synthetic (UH-S-cdiff1) images. A 10-fold cross validation is performed on the synthetic dataset and the result is reported with 95\% CI. }
\centering
\renewcommand{\arraystretch}{1.5}
\begin{tabular}{|l|l|c|c|c|c|c|c|}
\hline
\multirow{2}{*}{Dataset}     & \multirow{2}{*}{Method} & \multicolumn{2}{c|}{Vegetative Cell} & \multicolumn{2}{c|}{Spore} & \multicolumn{2}{c|}{Overall}  \\ 
\cline{3-8}
                             &                         & mAP       & Dice          & mAP       & Dice           & mAP       & Dice             \\ 
\cline{1-8}
\multirow{3}{*}{UH-A-cdiff1} & Mask RCNN               & 0.52      & \textbf{0.88}          & \textbf{0.69}      & \textbf{0.88}           & 0.54      & \textbf{0.88}             \\
\cline{2-8}
                             & FCRN                    & 0.41         & 0.60             & 0.13         & 0.46              & 0.23         & 0.60                \\
\cline{2-8}
                             & DETCID                  & \textbf{0.65}      & 0.85          & 0.65      & 0.87           & \textbf{0.65}      & 0.85             \\ 
\hline
\multirow{3}{*}{UH-S-cdiff1} & Mask RCNN               & 0.52 $\pm$ 0.03 & 0.86 $\pm$ 0.01     & 0.45 $\pm$ 0.05 & \textbf{0.90} $\pm$ 0.01       & 0.49 $\pm$ 0.02 & 0.83 $\pm$ 0.01        \\
\cline{2-8}
                             & FCRN                    & 0.36 $\pm$ 0.1         & 0.62 $\pm$ 0.03             & 0.46 $\pm$ 0.11         & 0.62 $\pm$ 0.03              & 0.25 $\pm$ 0.05         & 0.60 $\pm$ 0.03                \\
                             \cline{2-8}
                             & DETCID                  & \textbf{0.66 $\pm$ 0.02} & \textbf{0.88 $\pm$ 0.01}     & \textbf{0.69 $\pm$ 0.02} & 0.89 $\pm$ 0.01      & \textbf{0.67 $\pm$ 0.01} & \textbf{0.88 $\pm$ 0.01}    \\   
\hline
\end{tabular}
\label{table:performance}
\end{table*}

\subsection{Baseline comparisons}
The results were compared with two state-of-the-art methods in cell detection and segmentation:

\subsubsection{Mask-RCNN}
Mask-RCNN \cite{He_2017_ICCV} was developed by Facebook AI Research (FAIR) Lab as an instance-based object segmentation method capable of providing mask and bounding box for every object in the image. ResNet50 is used as the backbone with COCO pre-trained weights to be consistent with the backbone used for DETCID. A fully-connected region proposal network predicts rectangular bounding box regions. The regions of interest were resized and passed to the network head to compute the mask and refine the bounding box. 
We selected Mask-RCNN as a baseline since Mask-RCNN has achieved the state-of-the-art in instance-segmentation of cell nuclei in microscopy images \cite{feng2019robust} and segmentation of epithelial cells in pathology images with overlapping cells \cite{li2018path} using a TensorFlow implementation provided by Matterport which is used for comparison with DETCID \cite{matterport_maskrcnn_2017} and is initialized with COCO pretrained weights.

\begin{figure*}[bt]
  \begin{center}
\begin{tabular}{cc} \\

   \includegraphics[width=8cm,height=8cm]{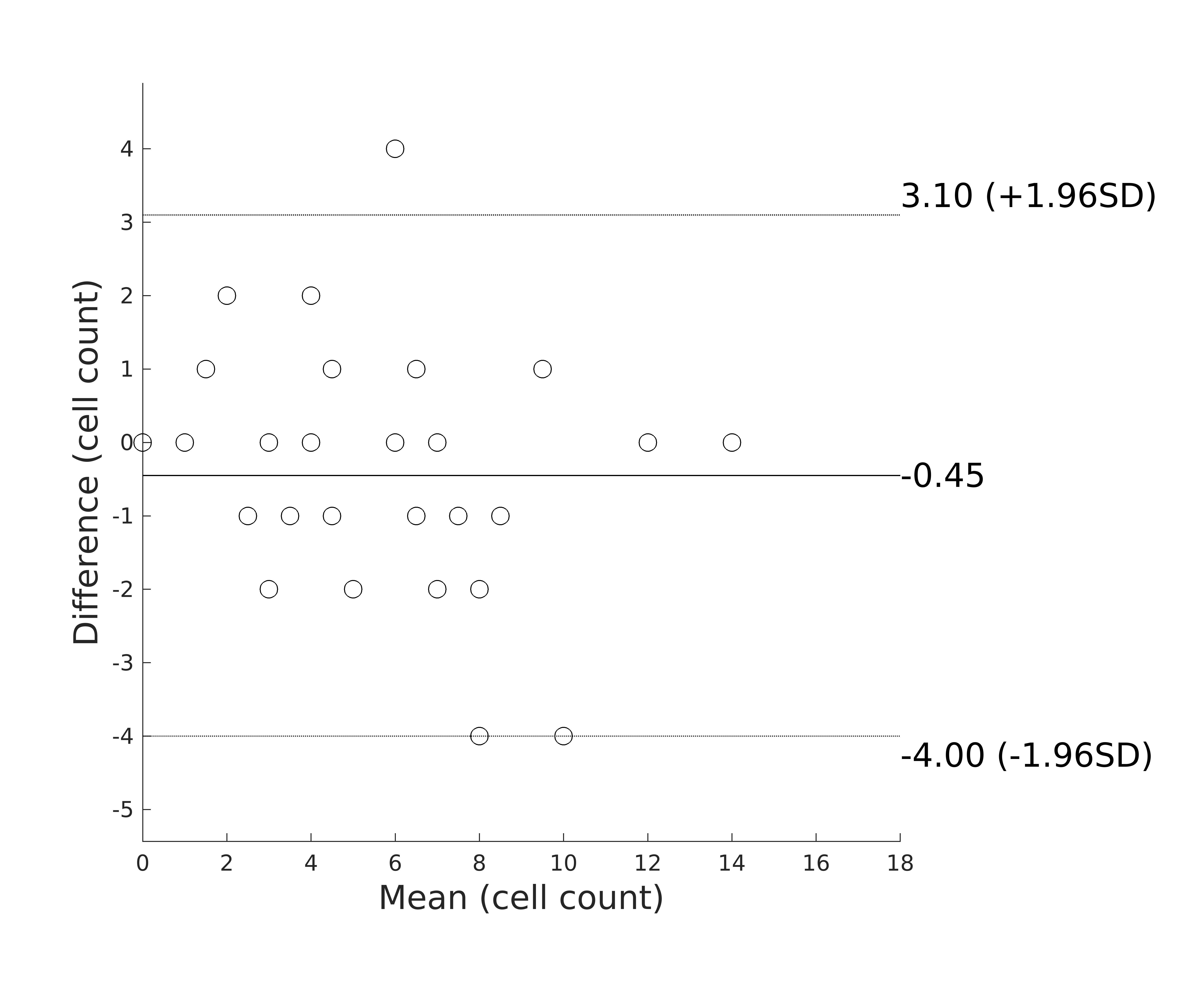}  &  
   \includegraphics[width=8cm,height=8cm]{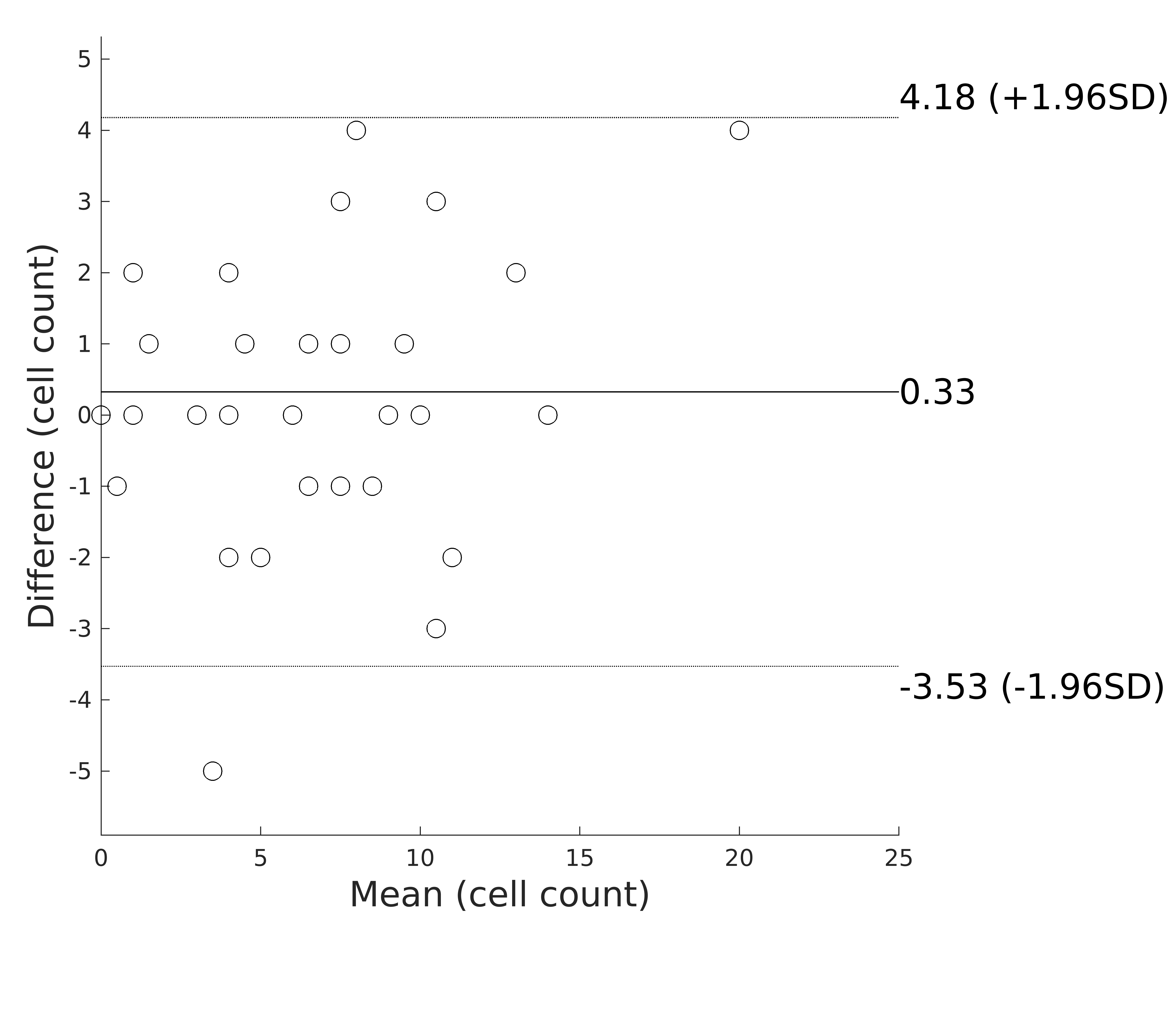}  \\ 
   (a) & (b) \\
   \includegraphics[width=8cm,height=8cm]{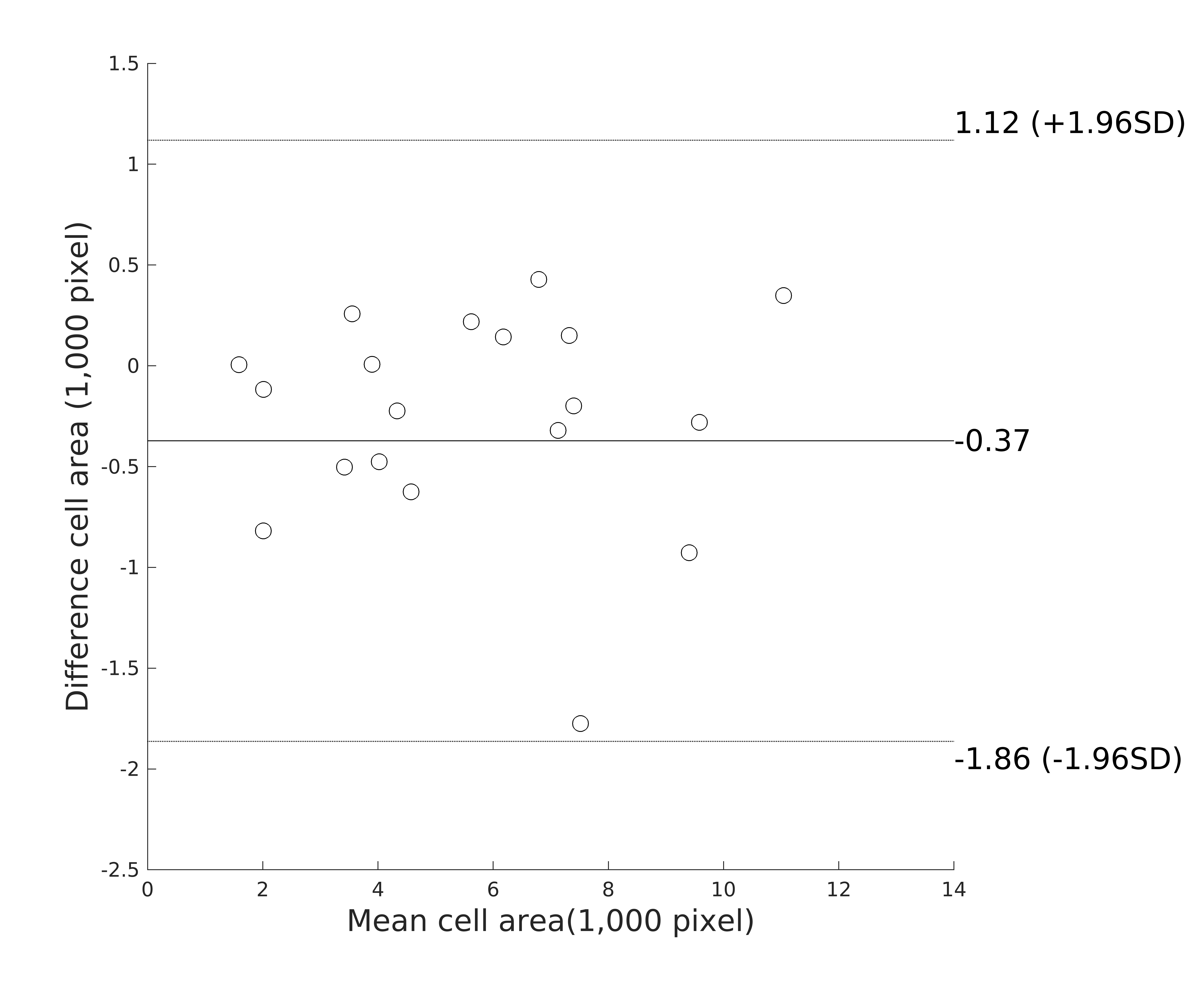}  &  
   \includegraphics[width=8cm,height=8cm]{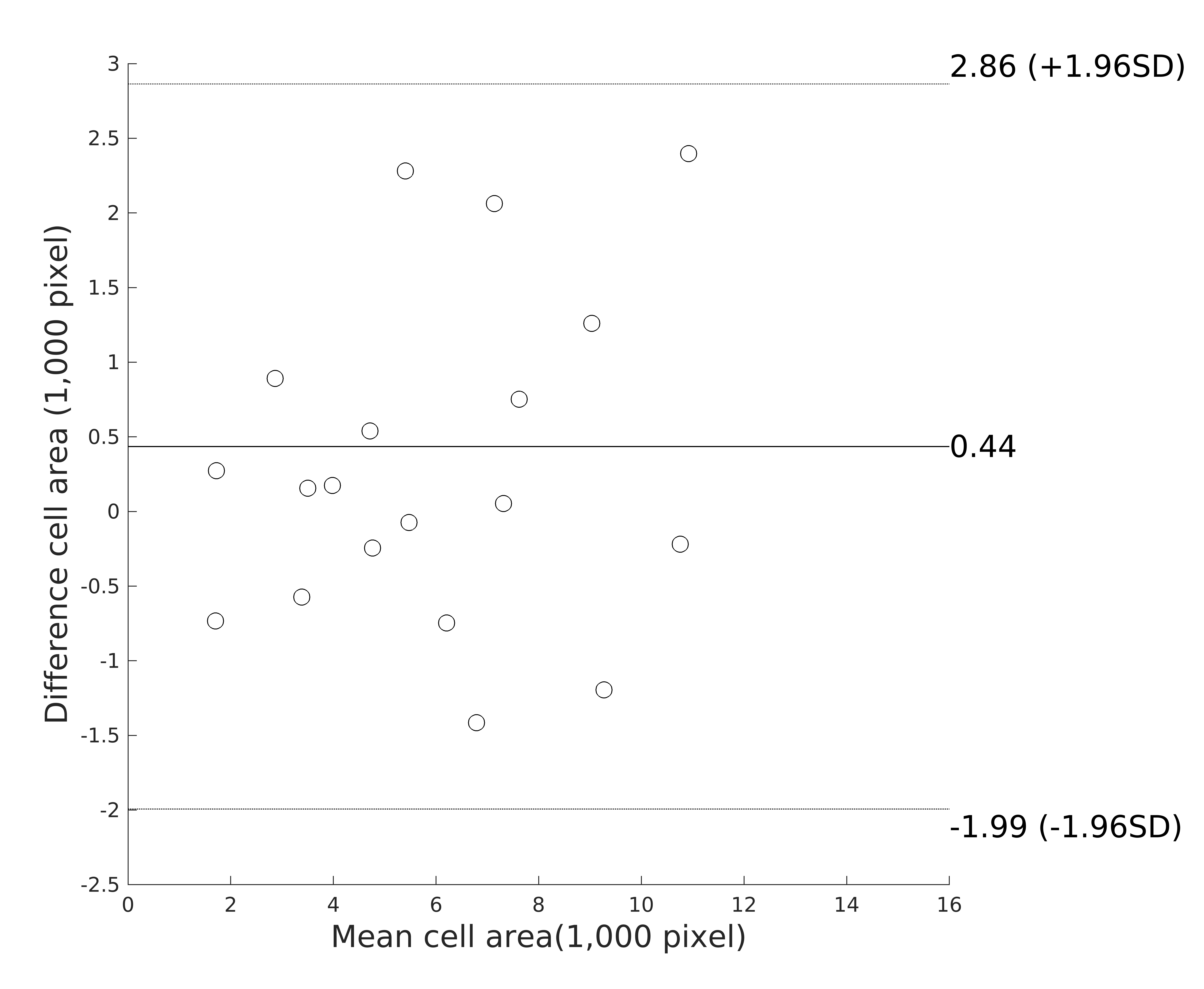}   
\\
(c) & (d)\\

\end{tabular}
\end{center}
  \caption{Depiction of Bland-Altman plots, comparing DETCID performance with the performance of the secondary set of human expert annotations: 
(a) the agreement between the two set of annotations on the number of annotated cells in image, 
(b) the agreement between the primary set of annotations and DETCID on the number of annotated cells in image,  
(c) the agreement between the two set of annotations on the annotated masks, and
(d) the agreement between the annotated in the primary set of annotations and the computed masks by DETCID.
}
\label{fig:BA_plot}
\end{figure*}

\subsubsection{Fully convolutional regression network (FCRN)}
Fully convolutional regression network (FCRN) is a U-net based cell detection method develop by Visual Geometry Group (VGG) at University of Oxford \cite{xie2018microscopy}. FCRN does not rely on a bounding box region of proposals and therefore is not sensitive to IoU threshold for detecting overlapping cells in various orientations. The network architecture includes a convolution and a deconvolution path predicting a density map for the image. The ground truth for every cell is defined as a 2D Gaussian where the pick is at the cell center. The label images were obtained by filtering the binary masks where the cell is white and the background is zero. The predicted local maxima in the image were found using Laplacian of Gaussian (LoG) and the watershed algorithm is used to detect the cell boundaries. We selected FCRN as a baseline because FCRN is capable of detecting cell clumps in various orientations using only synthetic training data.

\subsection{Experimental results}
UH-S-cdiff1 is used perform a 10-fold cross-validation to evaluate and compare the performance. Furthermore, UH-S-cdiff1 is used to train DETCID to evaluate on the acquired dataset UH-A-cdiff1. 
Mean average precision (mAP) and dice score were computed to measure the performance of the detection and segmentation respectively. 
Table \ref{table:performance} summarizes the quantitative performance evaluations. The results indicates that DETCID outperforms the state-of-the-art in detection of C. diff cells in the acquired SEM images UH-A-cdiff1 (\textit{{P=0.04; 95\% CI 0.001-Inf}}). However, DETCID achieves comparable results to the state-of-the-art in detection of spores in UH-A-cdiff1 (\textit{P=0.36; 95\% CI-0.11-0.28}) which is more challenging since they could be misclassified as debris due to their smaller size.
DETCID also achieved comparable results in segmentation of the vegetative cells  (\textit{P =0.63; 95\% CI -0.071-0.11}) and spores (\textit{P=0.15; 95\% CI -0.04-0.23}). 

Moreover, Table \ref{table:performance} indicates that DETCID achieved significant improvements in detection (\textit{P$<$0.001, 95\% CI -0.18-Inf}) and comparable results in segmentation (\textit{P=0.23; 95\% CI -0.03-0.13}) of UH-S-cdiff1 images where the number of touching clustered cells are higher.  
Therefore, DETCID outperforms the state-of-the-art in detection of C. diff cells where touching cells are clustered together. 

Figure \ref{fig:seg_result} depicts qualitative comparisons between DETCID performance and the state-of-the-art. The qualitative results indicate that Mask-RCNN has better performance in segmentation of isolated objects. However, DETCID outperforms Mask-RCNN when multiple cells are touching and in presence of debris resulting in lower mAP. FCRN is able to separate the touching cells. However, FCRN is highly sensitive to inhomogeneous illumination and presence of debris, resulting in poor mAP and dice score. 
 
\subsection{Bland–Altman analysis}
A Bland-Altman analysis is performed to compute the agreement between the result of DETCID and the primary manual annotation as well as the agreement two set of annotations for cell detection and segmentation. Figure \ref{fig:BA_plot} depicts the Blond-Altman plots for number of detected cells and the their segmentation masks. The Bland-Altman plots reveal no evidence of proportional bias for cell counts and segmentation differences. Furthermore, the number of detected cells by DETCID correlated with the primary set of annotations (${R^2=0.79}$) compared to the correlation between the two set of annotations ($R^2=0.81$). Accordingly, DETCID performance differs from the primary set of manual annotations in detection of CDI cells as the two set of manual annotations differ from themselves. The area of the segmentation masks computed by DETCID also correlated with the area of the annotated masks in the primary set (${R^2=0.89}$) compared to the area of the masks between the two set of annotations (${R^2=0.94}$).

\section{Conclusion}
In this paper, DETCID is proposed to detect and segment CDI cells in SEM images. An adversarial region proposal network was implemented to address the challenge of inhomogeneous illumination. Furthermore, a modified IoU metric is used for non-max suppression for detecting clusters of touching cells. A data augmentation algorithm was developed to provide a large number of training images suitable for training deep feature extraction architectures such as ResNet. The performance is compared to both deep region-based method and U-net based methods. DETCID outperforms the state-of-the-art in detection of touching cells and provide comparable result in segmentation of cells.

% ...

\bibliographystyle{IEEEtran}
\bibliography{DETCID}

\end{document}